\documentclass[conference,10pt]{IEEEtran_ICRATmod}

\IEEEoverridecommandlockouts
\usepackage[numbers]{natbib}

\usepackage{graphics} 
\usepackage{epsfig} 
\usepackage{times} 
\usepackage{amsmath} 
\usepackage{amssymb}  
\usepackage{amsthm}
\usepackage{amsfonts}
\usepackage{mathtools}
\usepackage{multirow} 
\usepackage{rotating} 
\usepackage{graphicx}
\usepackage{float}
\usepackage{flushend}
\usepackage{csquotes}
\usepackage{bm}
\usepackage{relsize}
\usepackage{epspdfconversion}
\usepackage{colortbl}
\usepackage{lipsum}
\usepackage[mathscr]{eucal}
\usepackage{fancyhdr}
\usepackage{comment}

\usepackage{tikz}

\usepackage{algorithm}
\usepackage[algo2e]{algorithm2e}

\usepackage{bbm}

\usepackage{hyperref}

\usepackage[many]{tcolorbox}
\usetikzlibrary{calc}
\usepackage[textsize=tiny]{todonotes}
\usepackage{xcolor}
\usepackage{graphicx}
\usepackage{tikz}
\usepackage[normalem]{ulem}

\definecolor{myblue}{RGB}{0,163,243}
\definecolor{mygreen}{RGB}{0,153,76}
\definecolor{mymagenta}{RGB}{190,0,190}

\tcbset{lhubstyle/.style={
		breakable,
		enhanced,
		outer arc=0pt,
		arc=0pt,
		colframe=myblue,
		colback=myblue!20,
		attach boxed title to top left,
		boxed title style={
			colback=myblue,
			outer arc=0pt,
			arc=0pt,
			top=3pt,
			bottom=3pt,
		},
		fonttitle=\sffamily
	}
}

\tcbset{mhubstyle/.style={
		breakable,
		enhanced,
		outer arc=0pt,
		arc=0pt,
		colframe=mygreen,
		colback=mygreen!20,
		attach boxed title to top left,
		boxed title style={
			colback=mygreen,
			outer arc=0pt,
			arc=0pt,
			top=3pt,
			bottom=3pt,
		},
		fonttitle=\sffamily
	}
}

\tcbset{shubstyle/.style={
		breakable,
		enhanced,
		outer arc=0pt,
		arc=0pt,
		colframe=mymagenta,
		colback=mymagenta!20,
		attach boxed title to top left,
		boxed title style={
			colback=mymagenta,
			outer arc=0pt,
			arc=0pt,
			top=3pt,
			bottom=3pt,
		},
		fonttitle=\sffamily
	}
}

\newtcolorbox[auto counter,number within=section]{lhub}[1][]{
	lhubstyle,
	title=\textbf{Response},
	overlay unbroken and first={
		\path
		let
		\p1=(title.north east),
		\p2=(frame.north east)
		in
		node[anchor=west,font=\sffamily,color=myblue,text width=\x2-\x1] 
		at (title.east) {#1};
	}
}

\newtcolorbox[auto counter,number within=section]{shub}[1][]{
	shubstyle,
	title=\textbf{Query},
	overlay unbroken and first={
		\path
		let
		\p1=(title.north east),
		\p2=(frame.north east)
		in
		node[anchor=west,font=\sffamily,color=mymagenta,text width=\x2-\x1] 
		at (title.east) {#1};
	}
}

\DeclareMathAlphabet\mathbfcal{OMS}{cmsy}{b}{n}

\def\compactify{\itemsep=0pt \topsep=0pt \partopsep=0pt \parsep=0pt}
\newcommand{\compress}{\itemsep=0pt \topsep=0pt \partopsep=0pt \parsep=0pt \leftmargin=30pt \labelwidth=30pt}

\let\latexusecounter=\usecounter

\begin{document}
\title{\textsc{ChatATC}: Large Language Model-Driven Conversational Agents for Supporting\\Strategic Air Traffic Flow Management}

\author{\scriptsize
	   	   \IEEEauthorblockN{Sinan Abdulhak, Max Z. Li}
	   	   \IEEEauthorblockA{University of Michigan\\Ann Arbor, MI, USA\\ \{\href{mailto:smabdul@umich.edu}{smabdul}, \href{mailto:maxzli@umich.edu}{maxzli}\}@umich.edu}
        \and
        \IEEEauthorblockN{Wayne Hubbard}
	   	   \IEEEauthorblockA{Federal Aviation Administration\\Warrenton, VA, USA\\ \href{mailto:Wayne.Hubbard@faa.gov}{Wayne.Hubbard@faa.gov}}
        \and
        \IEEEauthorblockN{Karthik Gopalakrishnan}
	   	   \IEEEauthorblockA{Stanford University\\Stanford, CA, USA\\ \href{mailto:karthikg@stanford.edu}{karthikg@stanford.edu}}
}


\vspace{-2cm}
\maketitle

\renewcommand{\headrulewidth}{0pt}
\lhead{\textcolor{black}{ICRAT 2024}}
\rhead{\textcolor{black}{Nanyang Technological University, Singapore}}
\cfoot{\textcolor{black}{\thepage}}
\lfoot{}
\thispagestyle{fancy}
\pagestyle{fancy}

\fancypagestyle{firststyle}
{
    \fancyhf{}
    \lhead{\textcolor{black}{ICRAT 2024}}
\rhead{\textcolor{black}{Nanyang Technological University, Singapore}}
    \cfoot{\textcolor{black}{\thepage}}
    \lfoot{\textcolor{white}{\thepage}\\ \vspace{0.1cm} \scriptsize \parbox{0.465\textwidth}{\hrule ~~\\ S. Abdulhak and M. Z. Li gratefully acknowledge software support from University of Michigan Information and Technology Services (UM ITS).}}
}

\noindent
\begin{abstract}
Generative artificial intelligence (AI) and large language models (LLMs) have gained rapid popularity through publicly available tools such as ChatGPT. The adoption of LLMs for personal and professional use is fueled by the natural interactions between human users and computer applications such as ChatGPT, along with powerful summarization and text generation capabilities. Given the widespread use of such generative AI tools, in this work we investigate how these tools can be deployed in a non-safety critical, strategic traffic flow management setting. Specifically, we train an LLM, \textsc{ChatATC}, based on a large historical data set of Ground Delay Program (GDP) issuances, spanning 2000-2023 and consisting of over 80,000 GDP implementations, revisions, and cancellations. We test the query and response capabilities of \textsc{ChatATC}, documenting successes (e.g., providing correct GDP rates, durations, and reason) and shortcomings (e.g,. superlative questions). We also detail the design of a graphical user interface for future users to interact and collaborate with the \textsc{ChatATC} conversational agent. 

\end{abstract}

\begin{small}{{\bfseries\itshape Keywords---Large language models; Traffic Management Initiative; Conversational agents; Decision support systems; Human-machine collaboration}}\end{small}

\thispagestyle{firststyle}
\section{Introduction}  \label{sec:intro}



Within the US National Airspace System (NAS), Traffic Managers must identify and respond to events that affect system demands and capacities. This strategic-level decision-making process requires consideration of, and development of appropriate responses to, a complex array of factors. Whether the traffic management issue is being evaluated by Traffic Managers at the level of individual Traffic Management Units, or across the entire NAS by National Traffic Management Officers, these positions demand a high level of expertise and experience \cite{FAA_responsibility}. Effective traffic management requires not only highly specialized training \cite{FAA_training}, but also continuous on-the-job learning. Additionally, as the NAS and the aviation industry move towards an increasingly data-driven, information-centric future \cite{FAA_icn}, decision support and automation systems should be developed to augment the aforementioned offline and experiential training.


As highly-specialized employees, Traffic Managers are also operating within a high-stress environment, a byproduct of the inherent complexities and difficult-to-predict nature of a large-scale infrastructure system such as the NAS. To this end, while unique challenges occur daily---sometimes hourly---which must be dealt with by Traffic Managers in collaboration with other NAS stakeholders, there are also issues which arise within the NAS that are more commonplace and repetitive. Operationally-impactful examples of the latter include, non-exhaustively, adverse wind conditions impacting arrival flows into New York TRACON (N90) \cite{stouffer2017}, fog-induced visibility limitations at San Francisco International Airport (SFO) \cite{mukherjee2012}, and (more recently) space launch operations affecting flows in Jacksonville (ZJX) and Miami (ZMA) Centers \cite{tinoco2021}. As a matter of workplace efficiency, it is desirable to allow Traffic Managers to focus on the former---unique challenges---employing dynamic and creative decision-making only possible through training and experiential learning. In contrast, formulating decisions and plans for routine NAS impacts need not start from a tabula rasa foundation: The question of \emph{how to enable such a process} is far more difficult. 

We give the following example: A Traffic Manager notes unfavorable wind conditions in the forecast for New York City. Simply handing over a historical data set containing decades of wind-induced Ground Delay Programs (GDPs) for N90 airports to this Traffic Manager, without any preprocessing, is unreasonable. However, preprocessing requires specifying, e.g., what GDP parameters (e.g., GDP rates, durations) might be of interest, or filtering across potentially dozens of other characteristics (e.g., GDP issuance reason, historic weather forecast). This nuanced data engineering process is critical for important tasks such as post hoc analyses and making long-term NAS investment decisions: However, from a workload perspective \cite{mannino2021} for an individual Traffic Manager, it is wholly inappropriate and out of scope, especially in the context of real-time NAS strategic traffic flow management decision-making.




\subsection{Generative AI and large language models}

Rapid advancements in generative artificial intelligence (AI) and the subsequent widespread adoption of generative AI-based technologies have the potential to fundamentally change the way people work \cite{NBERw31161}. Recently, the release and subsequent popularity of large language models (LLMs)---a type of generative AI which can provide general-purpose and human-like language generation (e.g., ChatGPT \cite{bubeck2023sparks})---showcased an example of this potential. In addition to language generation, such models are particularly adept at summarization and text synthesis \cite{grazil2021}. These powerful generative AI tools can greatly impact the productivity of new workers by accelerating training time or acting as a easy-to-use virtual assistant: According to the US National Bureau of Economic Research, generative AI-based conversational agents increased the number of issues resolved per hour by 34\%, based on a study of 5,179 customer support agents \cite{NBERw31161}. Such models can also be tailored to generate more specific responses: \emph{In-prompt learning} relies solely on the context provided by the user, whereas \emph{fine-tuning} a model uses additional input data during training.

Beyond customer service applications, given the potential of generative AI and LLMs to provide natural, \enquote{question and answer}-style interactions, such tools---if designed with the user in mind and deployed appropriately---offer a compelling, data-driven approach to enabling a more efficient process of dealing with repetitive, commonplace NAS issues. Revisiting the aforementioned example, the same Traffic Manager now types: \enquote{give a summary of past EWR\footnote{Newark Liberty International Airport; we refer to airports by their IATA code.} GDP examples due to wind,} and an LLM---trained or fine-tuned on text from GDP issuances and revisions---returns the information in a summarized, easy-to-read format. The Traffic Manager, cued by the LLM-driven conversational agent, no longer needs to approach this common NAS constraint from the ground up, potentially saving time and mental space for more unusual NAS challenges.

\subsection{Understanding user needs and tool limitations}  \label{sec:user_needs}

Leveraging co-author subject matter expertise, we first sought to understand GDP characteristics that are most important to Traffic Managers. We also honed in on specific ways that GDPs are represented, summarized, and displayed currently, which will inform the design and functionality of \textsc{ChatATC}. For Traffic Managers, the planning process typically begins the day before with PERTI meetings---\emph{Plan, Execute, Review, Train, and Improve} \cite{FAA_PERTI}---where potential airspace constraints are identified. If NAS constraints that could affect demand-capacity interactions are identified, Traffic Managers brainstorm potential traffic management solutions, of which GDPs are one possible course of action. Should circumstances warrant, a GDP can be proposed: In brief, GDPs seek to manage arrival demand into a capacity-constrained airport, proactively delaying flights on the ground at the origin to avoid costly airborne holding and potential diversion events \cite{FAA_TFMNASBooklet}. During day-of operations, Traffic Managers monitor real-time weather conditions as well as NAS demands (e.g., through the Flight Schedule Monitor, or FSM)---should NAS constraints develop as anticipated, the proposed GDP can be triggered. In case if the constraints are not as severe as previous forecasts had suggested, GDPs can be revised or canceled accordingly. 

Additional user-centric insights obtained prior to training \textsc{ChatATC} and proposing designs for the \textsc{ChatATC} user interface include:

\begin{itemize}
    \item Traffic Managers frequently rely on weather forecasts to determine if actions need to be taken, and these forecasts also inform the parameters of the proposed or implemented GDP;

    \item On top of the months of training required to understand the intricacies of GDP formulation and implementation, it often takes years of experience to ascertain optimal strategies that mitigate excessive flight delays while addressing demand-capacity imbalances;

    \item It will be critical to responsibly manage user expectations for a summarization tool such as \textsc{ChatATC}, for example, emphasizing that it is not a tool for prediction nor optimization; 

    \item There may be interest in using a tool such as \textsc{ChatATC} to query \emph{similar days}, as recalling similar events and how they were mitigated is time consuming and difficult for individual Traffic Managers.
\end{itemize}

\noindent
We note that on the last point regarding programmatically recalling similar days, a particularly salient future research direction may be to revisit the large body of literature revolving around identifying similar NAS days (see, e.g., \cite{gorripaty2017}). Such identification often utilized clustering-based and other quantitiative approaches: Could the strengths of previous identification methods be combined with LLM-driven capabilities to augment this similar days identification process? For example, one intersection could involve using LLMs (perhaps with in-prompt learning or fine-tuning, along with other modalities such as image processing \cite{GPT4_Visual}) to automatically generate descriptive, interpretable labels for similar NAS days identified through previous approaches.

We caveat that numerous limitations and challenges exist with LLMs: Three prominent risks include \emph{hallucinations}, wherein the LLM outputs unrealistic or implausible examples; \emph{verifiability}, as it can be difficult to quantify model performance and accuracy; and \emph{lack of interpretability}, as the decision-making process which resulted in a specific LLM output is almost entirely non-transparent. We note that ongoing research in AI and AI-related fields seeks to ameliorate such issues (e.g., chain-of-thought prompting to increase interpretability \cite{wei2023chainofthought}): However, these issues are not unique to the strategic air traffic management (ATM) setting---we emphasize that tools like \textsc{ChatATC} must be implemented in non-safety critical settings in light of these challenges.

\subsection{Research question and approach}

In this paper, we explore the use of generative AI and LLMs within the context of improving NAS situational awareness, reducing workload, and providing a powerful, information-centric tool that balances lack of specificity (e.g., providing generic information regarding GDPs) versus a deluge of information (e.g., providing every single GDP issuance without summaries or filters). Motivated by the popularization of LLMs and their potential to be appropriately adopted as virtual conversational agents to support Traffic Managers in easily accessing and gleaning insight from NAS operational data, we developed a research prototype called \textsc{ChatATC}. We focus on GDPs as the traffic management use case---\textsc{ChatATC} can rapidly retrieve, synthesize, and deliver summarized information regarding historical GDPs. Such summarizations may have been time-consuming or labor-intensive to access and create through traditional means. ATM users who may use tools similar to \textsc{ChatATC} include experienced Traffic Managers who want quick summaries of how previous NAS issues were handled, traffic management trainees interested in learning about common historical patterns of NAS issues, and adjacent personnel (e.g., FAA Air Traffic Organization Quality Control staff \cite{FAA_QAQC}) requiring contextual information regarding traffic management actions.

To highlight the need to appropriately deploy such a tool, we emphasize that \textsc{ChatATC} in no way supersedes the decision-making authority nor capabilities of the Traffic Manager. We remain mindful that such tools may have adverse impacts, e.g., in the ATM setting, \emph{over-reliance} \cite{choudhury2023} is of particular concern. Furthermore, we emphasize that \textsc{ChatATC} is not a prescriptive, optimization-like tool: It does not, and cannot, determine the \enquote{best solution} to a NAS issue.  






\subsection{Contributions of work}

To develop and train \textsc{ChatATC}, we first collected historical GDP data spanning approximately 23 years. Subsets of this data set containing historical GDP data is the main source of how we train and refine the LLM that drives \textsc{ChatATC}. Drawing on co-author subject matter expertise as a National Traffic Management Officer at the FAA Air Traffic Control System Command Center (ATCSCC, or Command Center), we examine the usefulness of \textsc{ChatATC} responses to a range of potential question inputs of interest to Traffic Managers. Finally, we contribute to the continued development of LLMs in support of ATM by designing an interface that Traffic Managers can use to interact with \textsc{ChatATC}.

\subsection{Manuscript organization}

The remainder of the manuscript is organized as follows: We provide a review of pertinent literature in Section \ref{sec:lit_review}, focusing on the impacts of generative AI and LLMs, as well as nascent research into the use of LLMs in aviation. 
We describe the historical GDP data set, introduce GDP data visualizations, and discuss various ways to train a GDP-specific LLM in Section \ref{sec:data_methods}. We detail our experiments and tests with \textsc{ChatATC} in Section \ref{sec:chatATC_testing} before concluding with a discussion of limitations and future work in Section \ref{sec:conclusion}.
\section{Literature review} \label{sec:lit_review}

There is extensive literature describing the potential role of generative AI technology in the workplace. \cite{woodruff2023knowledge} discuss worker expectations of generative AI’s impact based on a survey of 54 skilled workers and leaders across seven industries in the US, concluding that many see generative AI as complementing rather than replacing their work. Furthermore, \cite{noy2023} explore productivity effects of generative AI through a study of 444 college-educated professionals: Study participants were asked to complete an occupation-specific writing task. The authors found that study participants exposed to ChatGPT exhibited substantially improved productivity. Additional studies (e.g., \cite{felten2023language}) investigate the impact of generative AI on innovation, current on-the-job tasks, and wages. Divergent worker behavior regarding task division (delegation of tasks to the AI) versus task integration (continuous interaction with AI) were identified in \cite{dellacqua2023}---such studies re-emphasize the need to examine how individual Traffic Managers may interact with technology such as \textsc{ChatATC}. Finally, exploratory applications of LLMs such as ChatGPT in other specialized occupations include construction \cite{saka2024}, healthcare \cite{li2024_gpt}, emergency response \cite{Lovreglio2024}, and human resource management \cite{budhwar2023}.

\subsection{Large language models in aviation}

Although the application of LLMs to the aviation industry is in a nascent stage, there has already been a surge of research and prototyping interest. A powerful example of a general-purpose LLM specific to aviation is AviationGPT \cite{wang2023aviationgpt}, wherein the authors continuously train an LLM on unstructured text data pertinent to aviation operations (e.g., National Traffic Management Logs), though the focus is not specifically on GDP modeling, evaluation, and monitoring. Similar to our goal of mining GDP issuance data for training our LLM, \cite{tikayat2023} explores LLMs trained on aviation safety data via the Aviation Safety Reporting System (ASRS) data set, and \cite{ziakkas2023} examines historic aviation accidents. Zooming out beyond aviation to broader mobility and travel settings, LLMs have been used to extract emerging trends and synthesize potential research directions \cite{gursoy2023,klophaus2024}. Finally, \cite{wandelt2023} undertakes a comprehensive survey-based study examining the potential of LLMs in ATM from pedagogical and research perspectives. They found significant potential to increase learning efficiency and decrease time to lookup information, both of which are pertinent to ATM. The authors then ask ChatGPT for the best method to solve several combinatorial optimization problems: They found that while some of the methods and heuristics suggested by ChatGPT were a good starting point, others could occasionally be misleading.

Given that ChatGPT was first released to the public in November 2022, the less than two-year period (as of the time of writing) has seen the beginnings of a proliferation of initial research into how LLMs could be applied to aviation and mobility more broadly. Our work here contributes to this body of literature, examining a specific use case of LLMs for strategic ATM with a specific group of users in mind.
\section{Data and methods}  \label{sec:data_methods}

We set out to collect a large amount of historical data on GDP issuances via the FAA Operational Information System (OIS) website \cite{FAA_OIS_form}. Every time a GDP is proposed, created, modified, or cancelled, an entry is created on the OIS website, where data about the GDP issuance is stored in an XML format. For each entry, we extract the raw GDP text, and store this along with parsed parameters such as duration and associated Airport Arrival Rates (AARs). In total, we collected 86,842 GDP issuances spanning 146 airports between February 2000 and November 2023. 
We note that we were unable to locate GDP issuances prior to February 2000, and the end date of November 2023 was an artefact of when the data collection was performed. In practice, \textsc{ChatATC} should be periodically retrained (akin to the continuous update process described in \cite{wang2023aviationgpt}) as more GDP (and eventually, broadening out to other Traffic Management Initiatives and Traffic Flow Management actions such as Ground Stops, Airspace Flow Programs, Departure Sequencing Programs, Extended Metering activation, as well as Time-Based Flow Management) issuances occur.

\subsection{GDP summary statistics and program organization.}

\begin{figure}[!htbp] 
    \centering
    \includegraphics[width=0.75\linewidth]{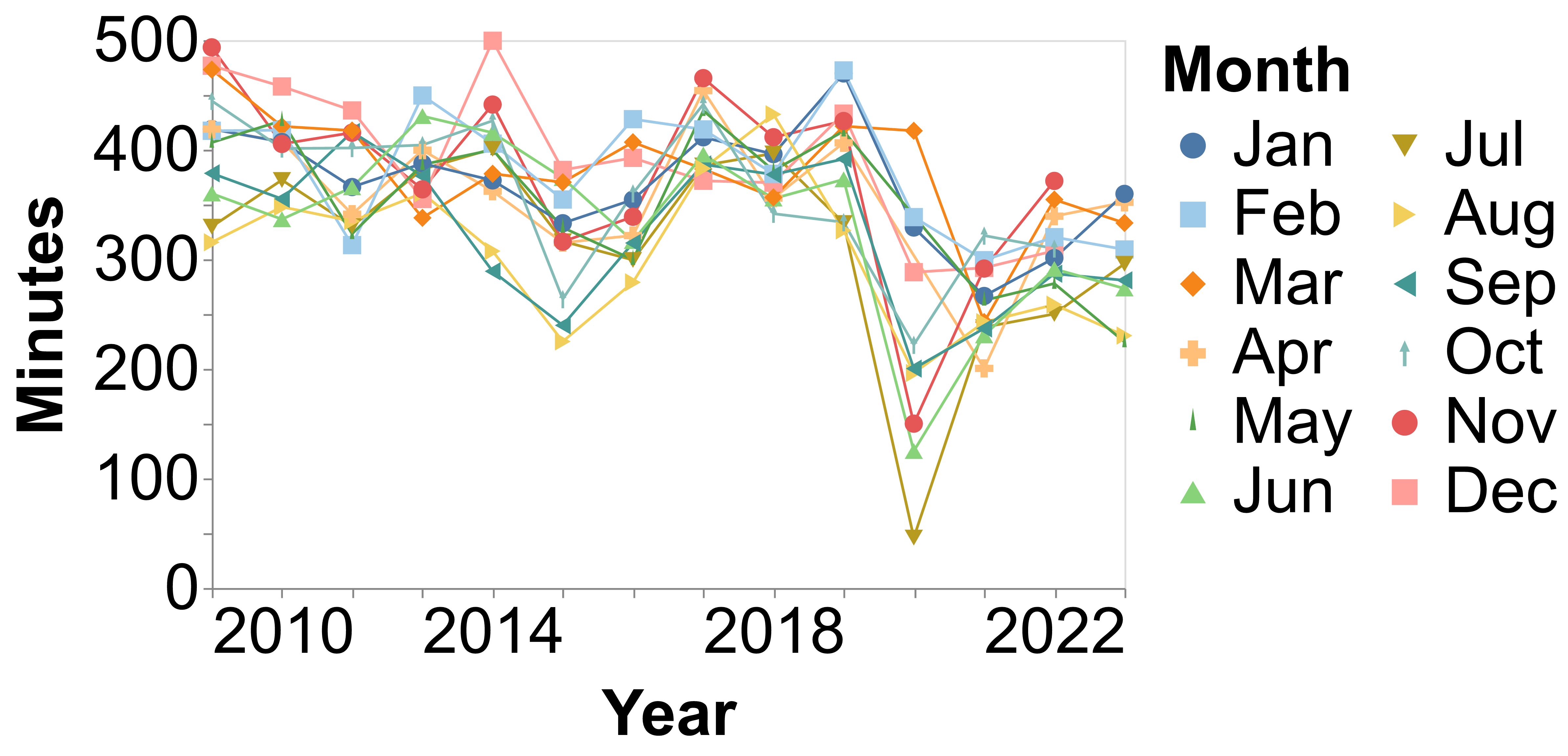}
    \caption{Average GDP duration from 2010 to 2023. Note that in July 2020 there was only one recorded GDP in the data set.}
    \label{fig:avg-gdp-duration}
\end{figure}

Prior to investigating training and fine-tuning an LLM using the collected GDP data, we provide summary statistics and data visualizations of our historical GDP data set. Note that visualizations span 2010-2023 due to differences in how GDP parameters were stored; however, \textsc{ChatATC} is trained on data from the entire date range (2000-2023). 
We first plot in Figure \ref{fig:avg-gdp-duration} average GDP duration over time to understand how this important parameter varied by month and year. We observe that winter months (e.g., November to February) typically have higher average GDP durations, likely due to severe weather and holiday traffic volume. Furthermore, a clear decline in average GDP duration during the COVID pandemic is visible in 2020.

\begin{figure}[!htbp] 
    \centering
    \includegraphics[width=0.8\linewidth]{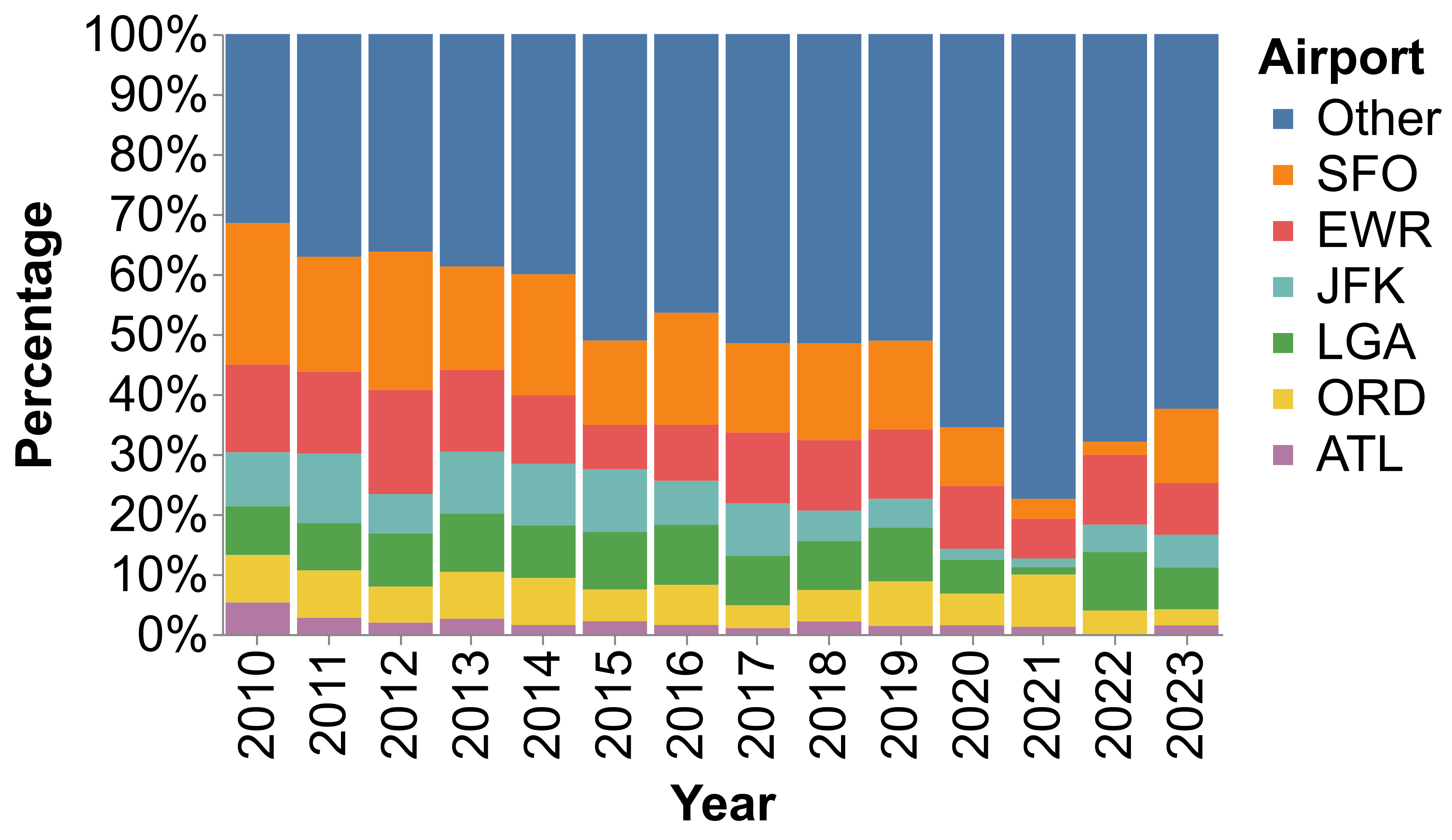}
    \caption{Percent of GDPs by airport from 2010 to 2023.}
    \label{fig:perc-gdp-fig}
\end{figure}

\begin{figure}[!htbp] 
    \centering
    \includegraphics[width=0.95\linewidth]{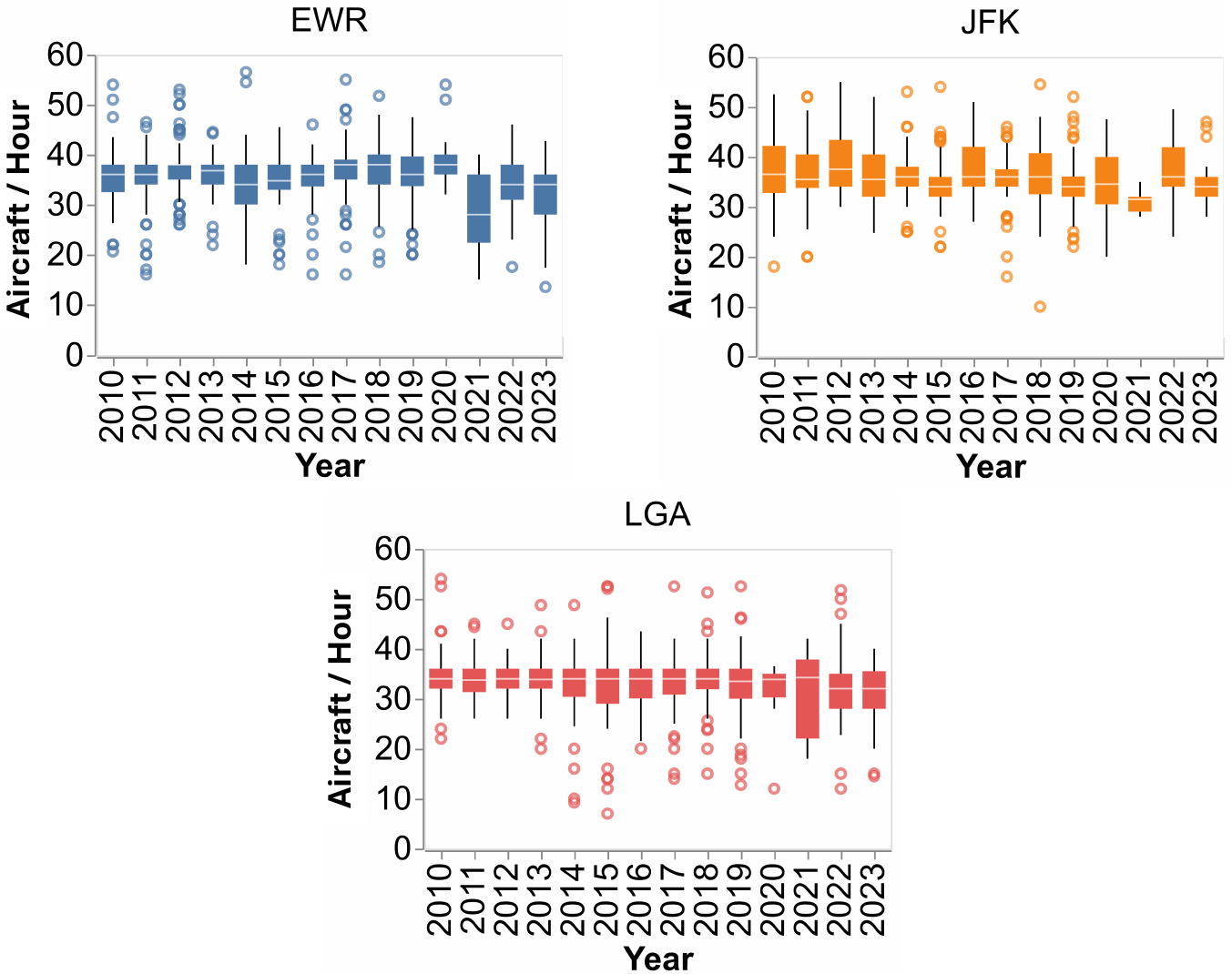}
    \caption{GDP rates from 2010 to 2023 for three major New York City airports. We assume GDP rates are nominally aircraft per hour; Traffic Managers may customize this to be, e.g., aircraft per 15 minutes, but this is rare.}
    \label{fig:ny-boxplot-fig}
    \vspace{-1cm}
\end{figure}

Next, we visualize what percentage of total GDPs come from 6 major airports across the US in Figure \ref{fig:perc-gdp-fig}. From this,  we identify that ORD has a higher incidence of GDPs as compared to ATL, and that post-2016, JFK has the lowest share of GDPs out of the major New York City airports. Furthermore, the share of GDPs that come from other airports steadily increases until 2021, where the trend reverses post-COVID. Lastly, for airport-level trends, we visualize the spread of GDP rates across the 3 major New York City airports in Figure \ref{fig:ny-boxplot-fig}.


\subsection{\textsc{ChatGPT}  summarization}

We explore two approaches to create a conversational agent with the ability to summarize information about historical GDPs: (1) use LLMs with in-prompt learning, or (2) LLMs fine tuned using the collected historical GDPs. We utilize two generative AI modeling tools: U-M GPT \cite{UMGPT} and Maizey \cite{UM_Maizey}. U-M GPT contains several LLMs that can be selected: GPT-3.5 (training data up-to-date as of September 2021), GPT-4 Turbo (training data up-to-date as of April 2023), and Llama 2 (training data up-to-date as of September 2022). In-prompt training for \textsc{ChatATC} relies only on U-M GPT, whereas fine-tuning will require Maizey---we describe Maizey later in this section. A significant differentiating factor between the services offered by U-M GPT versus commercially available LLMs such as ChatGPT is that U-M GPT is securely hosted internally at the University of Michigan. This alleviates potential concerns regarding researching LLMs while providing it potentially sensitive data. 
We first probe the effectiveness of in-prompt learning for \textsc{ChatATC}. 

\subsubsection{In-prompt or prompt-based learning}

U-M GPT enables experimentation with GPT-4 for the purposes of creating our ATM conversational agent. The most straightforward method for in-prompt learning would be to directly import the collected raw GDP texts as an initial prompt, and then proceed to interact with the LLM from the perspective of Traffic Managers. However, a major challenge with this simple approach is the constraints imposed on the length of input prompts that can be retained as a part of in-prompt learning. We could not directly import raw texts corresponding to the more than 80,000 GDP issuances and revisions, as this far exceeds in-prompt text capacities. As a workaround, we sort the GDP issuances by word count and paste the 13 wordiest issuances as a prompt into U-M GPT. After this, we experimented with a number of potential GDP-related inquiries, such as: \enquote{Tell me about the EWR GDP.} The output response is as follows:  


\begin{shub}[EWR GDP (in-prompt learning)]
	\enquote{tell me about the EWR gdp}
\end{shub}

\begin{lhub}[EWR GDP (in-prompt learning)]
	The ATCSCC advisory for Newark Liberty International Airport (EWR) issued on November 18, 2022, is   	proposing a Ground Delay Program (GDP) due to weather conditions, specifically wind. The delay was       calculated at 16:41 UTC (ADL time).

	$\quad$According to the advisory, the arrivals estimated for this delay lay between 18:00 UTC on the 18th 	 
	of November to 03:59 UTC on the 19th of November. 

	$\quad$During this period, the anticipated program rate is given as a repeating sequence: 34/34/34/38/38/38/38/\\38/38/38. 

	$\quad$The delay is applicable for all departures from contiguous US and several depature points from Canada.

	$\quad$The estimated maximum delay time is 105 minutes and the average delay time is 44 minutes. 

	$\quad$The message also mentions data issues prior to 21:00 UTC and a shift to DAS GDP mode to avoid possible delay limit in a UDP mode. 

	$\quad$Additionally, user updates must be received by 17:05 UTC on the 18th of November. The effective time of the advisory is from 16:44 UTC on the 18th of November to 17:59 UTC on the same day. 
\end{lhub}

Even with only a shortened version of in-prompt learning, the LLM is able to recognize the general structure of a GDP issuance, and accurately extract parameters from the GDP issuance with no additional prompting or guidance. These parameters include the GDP rate, affected departure airports (i.e., the scope of the GDP), among others. We manually checked the raw text of the 13 GDP entries used for in-prompt learning to ensure that the LLM is reporting accurate information. There has been nascent work that explores more rigorous in-prompt learning to reduce hallucinatory responses \cite{jha2023_inprompt}---future work in the strategic ATM setting could investigate how to specify instructions (in prompt) to the LLM that best mitigates hallucinations when it comes to GDP summarizations. 

\begin{shub}[GDP due to weather (fine-tuned model)]
	\enquote{give me an example of gdp due to weather}
\end{shub}

\begin{lhub}[GDP due to weather (fine-tuned model)]
	One example of a Ground Delay Program (GDP) due to weather is the advisory from 18/11/15. The message indicates that there is a proposed ground delay at SFO with an anticipated maximum delay of 64 minutes and an average delay of 35.8 minutes. The reason given for the delay is weather, specifically low ceilings.
\end{lhub}

\subsubsection{Fine tuning LLMs}

Maizey is a generative AI tool that allows for fine-tuning LLMs via a pre-constructed repository of text inputs. For example, this could take the form of a directory with individual text files and documents of interest. We use Maizey to fine tune GPT-4---using the large set of collected GDPs---into a custom LLM, \textsc{ChatATC}, and explore its response capabilities to a variety of inputs that Traffic Managers may be interested in. For the fine-tuning process, we provided the raw text files of collected GDP issuances (in a combined \texttt{.txt} format) to Maizey. 

Based on guidance regarding best practices for using Maizey~\cite{UM_Maizey}, along with considering how Traffic Managers might use \textsc{ChatATC} in real-world settings, we decided to train separate instances of \textsc{ChatATC} conversational agents for specific airports. This also allows us to train the model with GDPs for a specific airport for more relevant and context-rich responses. To reduce the computational burden of fine-tuning the LLM, for each instance of \textsc{ChatATC}, we select 500 historical GDP observations that contain the highest amount of text. Doing so provides us with a range of historical GDP issuances with different parameters settings. For our experiments described in later sections, we created instances of \textsc{ChatATC} for SFO and EWR, two large US airports that often contend with GDPs \cite{mukherjee2012,stouffer2017}. Finally, for each instance of \textsc{ChatATC}, during the process of fine-tuning, we also adjust the \texttt{temperature} hyperparameter within Maizey: This parameter controls for output sensitivity and randomness, and can be interpreted as way to elicit more creative responses from the LLM \cite{UM_Maizey}.

\section{Testing out \textsc{ChatATC}}  \label{sec:chatATC_testing}


As we discuss the outputs from \textsc{ChatATC}, we re-emphasize that safety is critical in the aviation domain: We envision \textsc{ChatATC} being applied only in non-safety critical roles, given the risks of hallucinations, lack of clarity on how to verify its outputs, and lack of interpretability. To increase our likelihood of obtaining factually correct outputs, we chose a \texttt{temperature} parameter of 0.2 in Maizey for our test cases and experiments (lower parameter values result in \enquote{less creative} answer ouputs). We note, however, that we did not find a significant difference in the accuracy of \textsc{ChatATC} responses for \texttt{temperature} parameter settings between 0.2 and 1.2, perhaps due to the specificity of the queries in this ATM use case; additional testing on this front is needed.

After fine tuning \textsc{ChatATC} via Maizey, we started by asking the SFO instance of \textsc{ChatATC} about GDPs due to weather. We observed that \textsc{ChatATC} was able to quickly retrieve an example of a relevant GDP from one of the 500 historical GDP issuances it had been fine tuned on. Specifically, the SFO instance of \textsc{ChatATC} was able to retrieve examples of GDPs on a specific date, GDPs due to a specific impacting condition (e.g., poor weather), GDPs with a specific rate, among others. However, we observed limitations when it comes to superlatives-based questions. For example, asking \textsc{ChatATC} for the GDP with the \emph{highest maximum delay} resulted in an incorrect answer. However, we note that Traffic Managers are less likely to ask superlatives-based questions, as compared with questions regarding, e.g., the rate of a previous GDP, or the reason why a GDP was implemented---we do observe factually correct answers to the latter questions.  


\begin{shub}[GDP superlatives]
	\enquote{which gdp at sfo had the highest max delay}
\end{shub}

\begin{lhub}[GDP superlatives (incorrect response)]
	The GDP at SFO that had the highest max delay was the one mentioned in the message with RAW TEXT: ATSCC ADVZY 045 SFO/ZOA 02/16/2009 CDM GROUND DELAY PROGRAM with a maximum delay of 784.

    \textcolor{red}{[A manual search through SFO GDP reveals maximum delays of 1444 minutes, not 784.]}
\end{lhub}

Because the rate, start and end times, and prevailing runway configuration are particularly important parameters to users of a conversational agent such as \textsc{ChatATC}, we relied on prompt engineering by adding the following to \textsc{ChatATC}'s system prompt: \enquote{Give me date, start time, end time, program rate, runway configuration and impacting condition.} Furthermore, we wanted the output from \textsc{ChatATC} to be more structured: We included the following system prompt as well: \enquote{Give your response in a bulleted list.} After these two system prompts, \textsc{ChatATC} provided outputs that were more consistently formatted, and can be easily parsed by the user. Additionally, the ability to pre-specify output formats is critical for connecting \textsc{ChatATC} with an easy-to-use graphical user interface (GUI).

 

\subsection{Graphical user interface design}


Future interactions with conversational agents such as \textsc{ChatATC} can be greatly enhanced through a well-designed front end, i.e., a GUI. The basic functionalities of this GUI must include the ability for users to type in an input query, and display the response from \textsc{ChatATC} in a logical manner (e.g., text-only, graphics-only, or mixed modality, depending on the query and response). To this end, we develop a wireframe for such a front end, visualizing how \textsc{ChatATC}'s responses may be displayed, with particular emphasis on highlighting parameters that are important to Traffic Managers (e.g., program rates). We also paid particular attention to the hierarchy and order of information presented, placing identifying information such as airport, impacting condition, and start/end times at the very top of the screen. A well-organized GUI could help users of \textsc{ChatATC} rapidly identify and retrieve the requested information pertaining to GDPs (or broadly, any TMIs). 

\begin{figure}[!htbp] 
    \centering
    \fbox{\includegraphics[width=0.975\linewidth]{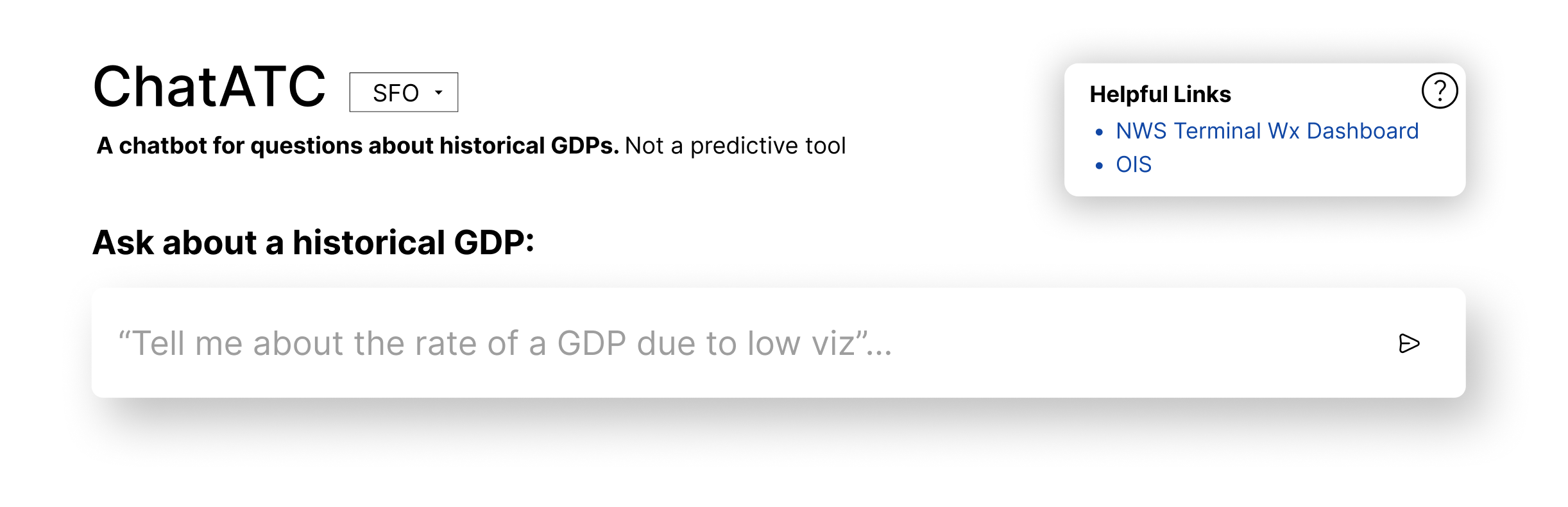}}
    \caption{Home page for \textsc{ChatATC} in GUI wireframe.}
    \label{fig:homescreen_topscreen}
\end{figure}

\begin{figure}[!htbp] 
    \centering
    \fbox{\includegraphics[width=0.975\linewidth]{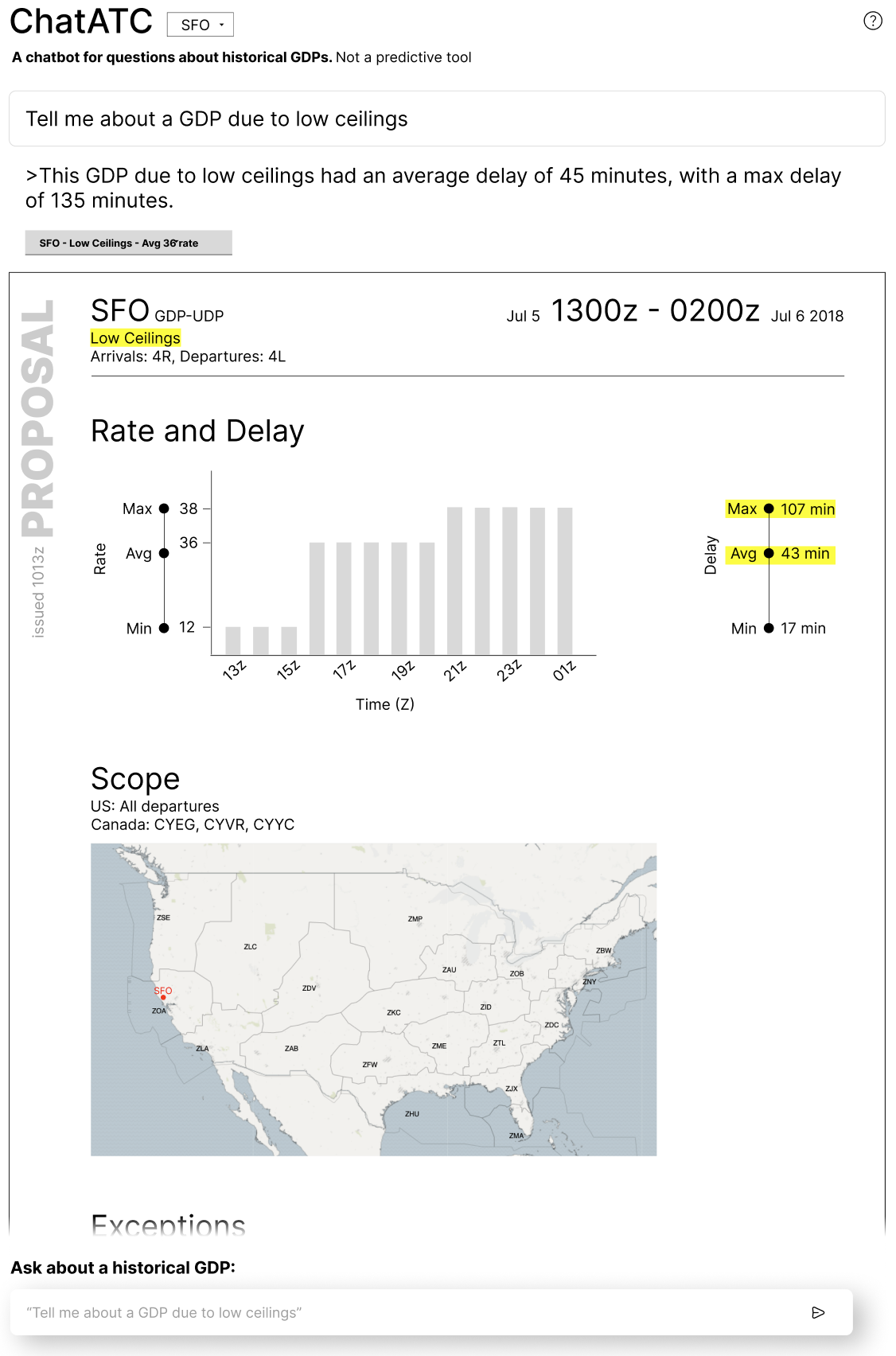}}
    \caption{GUI output for \textsc{ChatATC} query with GDP parameters.}
    \label{fig:final_iteration}
\end{figure}

We sought to minimize the number of choices users have to make when accessing the \textsc{ChatATC} GUI for the first time. On the home screen, there is a single drop-down menu where users can select the desired instance of \textsc{ChatATC} (e.g., the instance trained on SFO GDPs versus EWR GDPs). A large field for the user to enter a query spans the entire width of the screen and directs the user's attention to it. To manage user expectations, a short description is included explicitly on the home page (Figure~\ref{fig:homescreen_topscreen}) which states that \textsc{ChatATC} is not a predictive tool, emphasizing the non-safety critical nature of this conversational agent. As we iteratively improved the GUI design, we implemented a direct link to the National Weather Service Terminal Weather Dashboard, as this is a tool that is often referenced by Traffic Managers. Users now can easily reference weather- and convective activity-related information while using \textsc{ChatATC}. In addition to this, we also provide a direct link to relevant FAA OIS pages. The final iteration can be seen in Figures \ref{fig:homescreen_topscreen} and \ref{fig:final_iteration}, the latter of which depicts how we chose to visualize the GDP rate, delay metrics, and GDP scope. Ongoing and future research will investigate dynamic GUI outputs which are responsive to the specific types of query and \textsc{ChatATC} outputs, showing users different GUI themes that best capture the requested information.

\section{Conclusion and future work}    \label{sec:conclusion}

We investigate the use of generative AI models to assist NAS Traffic Managers in summarizing historical GDP information, sifting through loosely organized GDP data that would otherwise be laborious to filter and extract pertinent information from. Such tools can help existing Traffic Managers with future ATM decision-making or help increase the quality of training for new personnel. We collected over two decades of historical GDP data, amounting to over 80,000 GDP issuances (new GDPs, revisions to GDPs, and GDP cancellations). We used this set of historical GDP data to train \textsc{ChatGPT}, an LLM built on GPT-4, to answer questions with natural responses about historical GDPs. We also designed a user interface to enable interactions and future collaborations between Traffic Managers and \textsc{ChatATC}. We believe the simplicity and accessibility of such a tool will be important for adoption, particularly since its application is in a non-safety critical area (i.e., strategic traffic flow management). 

We elaborate on the limitations of our work, along with future research directions: While the current performance of \textsc{ChatATC} is promising, there are aspects of its responses that can be improved. For example, since incorporating prompt augmentation to create more detailed and consistent outputs, \textsc{ChatATC} has been unable to retrieve exact GDP dates, despite this information being readily available in the raw GDP text. A major effort for future research is the need to establish a baseline by which to compare and evaluate the performance of large language models in aviation. Another future exploration could be experimenting with training \textsc{ChatATC} using raw GDP text with different file formats, or experimenting with further prompt augmentation to nudge \textsc{ChatATC} about the location of dates in the raw GDP text. Additional user testing and feedback collection need to be performed with a larger group of Traffic Managers, then expanding out to other potential users.

\section*{Disclaimer} \label{sec:ack}

\noindent
The contents of this document reflect the views of the authors and do not necessarily reflect the views of the Federal Aviation Administration (FAA) or the Department of Transportation (DOT). Neither the FAA nor the DOT make any warranty or guarantee, expressed or implied, concerning the content or accuracy of these views.

\bibliographystyle{IEEEtran} 
\small{
\bibliography{main.bib}
}


\end{document}